\documentclass[10pt,conference]{IEEEtran}
\usepackage{times,amsmath,epsfig,multirow,tabularx}
\usepackage{epstopdf}
\usepackage{graphicx,amsmath, amsfonts, amssymb,mathtools}
\usepackage[caption=false]{subfig}

\newcommand\clearrow{\global\let\rowmac\relax}
\clearrow

\title{Can you tell a face from a HEVC bitstream?}
\author{%
{Saeed Ranjbar Alvar, Hyomin Choi and Ivan V. Baji\'c} 
\vspace{1.6mm}\\
\fontsize{10}{10}\selectfont
School of Engineering Science, Simon Fraser University, Burnaby, BC, Canada \\ 
Email: \{saeedr,chyomin, ibajic\}@sfu.ca
\,\\ 
\\
\fontsize{9}{9}\selectfont\ttfamily\upshape
\vspace{1.2mm}\\
\fontsize{10}{10}\selectfont\rmfamily\itshape
\,\\ 
\\

\fontsize{9}{9}\selectfont\ttfamily\upshape
\,
}
\begin{document}
\maketitle



\begin{abstract}
Image and video analytics are being increasingly used on a massive scale. Not only is the amount of data growing, but the complexity of the data processing pipelines is also increasing, thereby exacerbating the problem. It is becoming increasingly important to save computational resources wherever possible. We focus on one of the poster problems of visual analytics -- face detection -- and approach the issue of reducing the computation by asking: Is it possible to detect a face without full image reconstruction from the High Efficiency Video Coding (HEVC) bitstream? We demonstrate that this is indeed possible, with accuracy comparable to conventional face detection, by training a Convolutional Neural Network on the output of the HEVC entropy decoder.
\\   
\end{abstract}


\section{Introduction}
Faces are important for visual analytics. The availability of large datasets containing images with faces from various social platforms, combined with the emergence of advanced machine learning architectures such as deep neural networks, have led to fairly reliable face detection and localization capabilities. In this paper, we use the term \emph{detection} (e.g., face detection) to mean deciding on the presence of the object (face) in an image, as is common in detection theory~\cite{Kay_Detection}. Meanwhile, finding the location of the face in the image will be referred to as \emph{localization}. It should be noted that in some recent literature, the term \emph{detection} has been used to imply both detection and localization. 

Recent techniques such as~\cite{cascade, facial_parts, hyperface} use large datasets to train deep convolutional neural networks (CNN) to detect and/or localize faces. Through the training process, the weights of the CNNs used in these methods are adjusted to find features that can effectively differentiate between the face and non-face image patches. This capability lays the foundation for subsequent stages of the processing pipeline, such as counting the number of people in the scene, extracting facial features and landmarks, and so on. 

On the other hand, visual analytics, especially those involving CNNs, are computationally expensive~\cite{He2014_arXiv}. Given today's massive scale of visual data on which analysis is supposed to run, it is becoming imperative to save the computational effort wherever possible. When it comes to computer vision, one of the often overlooked computational bottlenecks is image/video decoding, especially with the most recent video coding standard HEVC~\cite{HEVC}. So, in this work, we set out to examine whether it is possible to detect a face in a HEVC bitstream without full decoding and image reconstruction. Specifically, we look at the output of the HEVC entropy decoder in intra-coded images. As shown in Table~\ref{Tbl:complexity}, HEVC entropy decoder takes, on average, less than 40\% of the overall decoding time, depending on the resolution. In this table, different sequence classes correspond to different resolutions~\cite{hevc_ctc}. Using the output of the HEVC entropy decoder, we train a simple shallow CNN to detect faces based on several HEVC syntax features. This strategy turns out to be as effective as conventional face detection in a fully decoded image, but with lower computational cost.  

\begin{table}[t]
	\centering
	\caption{HEVC entropy decoding time as a percentage of full decoding and reconstruction in the all-intra mode}
	\label{Tbl:complexity}
\begin{tabular}{|c|c|}
	\hline
	Sequences & \% time \\ \hline
	\hline
	Class A & 35 \\ \hline
	Class B & 37 \\ \hline
	Class C & 36 \\ \hline
	Class D & 43 \\ \hline
	Class E & 30 \\ \hline
	\hline
	Average & 37 \\ \hline
\end{tabular}
\vspace{-0.3cm}
\end{table}

While compressed-domain analytics has been studied for a number of years, there has been very limited work on visual analysis using HEVC bitstreams. The few examples include~\cite{hevc_counting}, where the number of moving objects in the scene is estimated without full HEVC decoding, and~\cite{moving_object_hevc}, where moving objects in surveillance video are classified into humans or vehicles. The present work adds to this recent body of literature on HEVC-domain analytics and, to our knowledge, is the first work on face detection in HEVC bitstreams.  
The paper is organized as follows. Section~\ref{sec:proposed_method} presents the details of the proposed face detection approach. Section~\ref{sec:Exp_results} discusses the experimental results, followed by conclusions in Section~\ref{sec:Conclusion}.

\section{Proposed Method}
\label{sec:proposed_method}
\begin{figure}[t]
	\centering		
	\begin{subfloat}{}
		\centering
		\centerline{\includegraphics[scale=0.35]{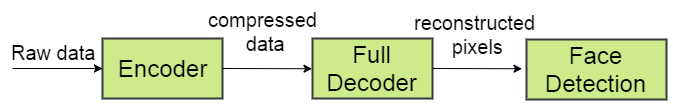}} {(a)} \\
		\label{fig:PD}
	\end{subfloat}
	
	\begin{subfloat}{}	
		\centerline{\includegraphics[scale=0.35]{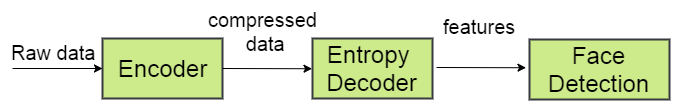}} {(b)} \\
		\label{fig:CD}	
	\end{subfloat}
	 	\caption{(a) Conventional face detection; (b) proposed face detection.}
	 	\label{fig:flow_chart}
\end{figure}

Unlike conventional face detection (Fig.~\ref{fig:flow_chart}(a)), the proposed detection system operates on the features obtained from the HEVC intra-coded bitstream, at the output of the HEVC entropy decoder (Fig.~\ref{fig:flow_chart}(b)). By skipping subsequent HEVC decoding stages -- dequantization, inverse transformation, prediction and pixel reconstruction -- significant savings in decoding time can be achieved, as indicated in Table~\ref{Tbl:complexity}. The proposed system is intended to operate on HEVC-coded still images or I-frames. For inter-coded frames, one can adopt a motion vector-based tracking method such as~\cite{moving_object_hevc, kb_tip_2013} to follow the faces in between the I-frames.   



The operation of the proposed face detector is explained using square-shaped patches. In the explanation as well as experiments we use patches of size $64 \times 64$ and $128 \times 128$ as examples, but the main ideas apply to other patch sizes as well. Based on the data at the output of the HEVC entropy decoder, we construct three \emph{feature channels} of the same size as the patch, and combine them in a 3-channel \emph{feature image}. This feature image is fed to a CNN whose task is to decide whether or not the patch contains a face. 


\subsection{Creating the Feature Image}
\begin{figure}[t]	
	\centering
	\centerline{\includegraphics[scale=0.3]{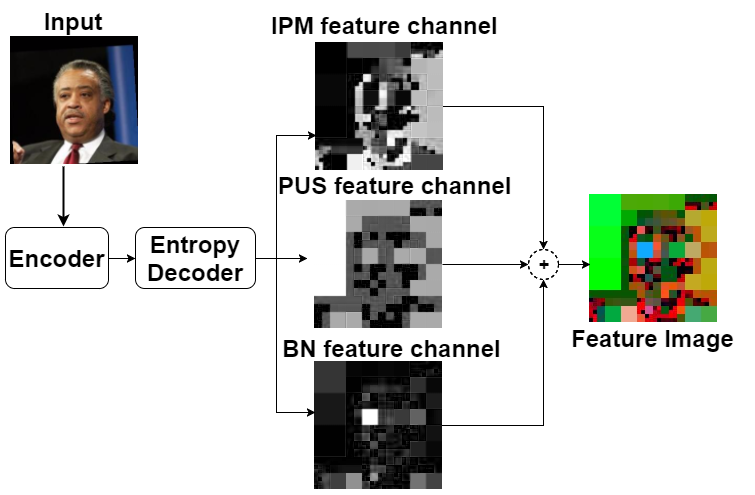}}
	\caption{Creating the feature image.} 
	\label{fig:feature_image}
\end{figure}

During HEVC entropy decoding, the Intra Prediction Mode (IPM), Prediction Unit Size (PUS) and Bin Number (BN) are reported for each Prediction Unit (PU). We map these values to a range 0-255 and then copy them into the corresponding location in the image, as shown in Fig.~\ref{fig:feature_image}.  

IPM values, which are numbers in the range 0-34, are linearly mapped (and rounded when needed) to integers 0-255. PUS values can be $\{4, 8, 16, 32\}$ and they are mapped to $\{0, 85, 170, 255\}$. 
Finally, BNs in each PU vary depending on bit consumption, which in turn depends on the complexity of the underlying signal. We first find the minimum and maximum BN in the image and then linearly map that range to 0-255, rounding the result when necessary. This way, we create a feature image  that can be visualized and processed in a similar way to conventional 3-channel images. 


Since the smallest coding unit in HEVC is $4 \times 4$ and each feature value is reported once per coding unit, the feature channels (as well as the final feature image) could be $\frac{1}{16}$ of the size of the input patch. However, we decided to extend the feature channels and the feature image to the full size of the input patch to facilitate easier visualization and comparison with pixel-domain face detection. 

\begin{figure}[t]	
	\centering
	\centerline{\includegraphics[scale=0.3]{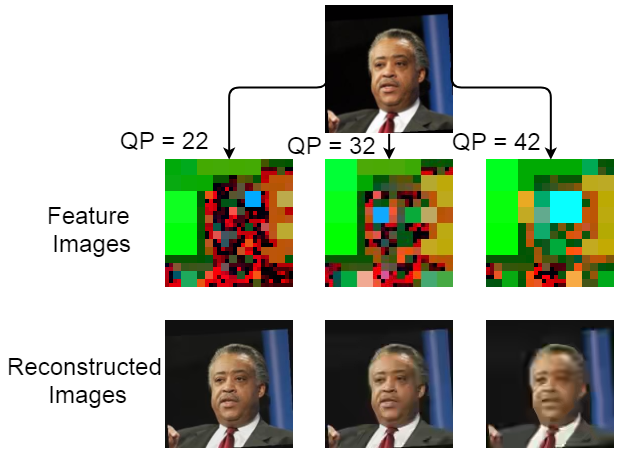}}
	\caption{An example of feature images and fully reconstructed images for the input encoded with three different QP values.} 
	\label{fig:QP_effect}
\end{figure}

Note that feature images change when the Quantization Parameter (QP) value changes, because encoding decisions regarding prediction modes, PU sizes, etc., all depend on QP. Fig.~\ref{fig:QP_effect} shows an example, where an image patch containing a face is encoded with QP $\in \{22, 32, 42\}$. As seen in the figure, feature images change more than the resulting fully-reconstructed images shown in the bottom of the figure. Hence, one could expect that face detection from feature images may be more challenging than conventional pixel-domain detection. 

\subsection{CNN for Face Detection from HEVC Feature Images}
\label{subsec:Train_CNN}

The CNN architecture for face detection from HEVC feature images was selected in the following way. We encoded a number of $128 \times 128$ image patches with and without faces using QP = 32 and constructed feature images from them. Our dataset is explained in more detail in Section~\ref{sec:Exp_results}. 

We started with a very simple network (implemented in Keras\footnote{https://keras.io/} with Tensorflow backend) comprising one convolutional layer (with one $5 \times 5 \times 3$ filter, stride of 4) and one fully-connected layer with one unit connected to the output, whose value is used for face/non-face decisions. The CNN was trained using Stochastic Gradient Descent (SGD) with the learning rate of $10^{-4}$. We started increasing the number of units in the fully connected layer and stopped at 500, where the accuracy saturated. Then we started to increase the number of filters in the convolutional layer and observed that the accuracy kept increasing until the number of filters reached 100, where it saturated. At this point we added the max-pooling layer with window size of $2 \times 2$ and stride of 2 to the convolutional layer (which improved the accuracy) and then added the second convolutional layer with 100 filters. The accuracy did not improve beyond the 100 filters in the second convolutional layer, so we kept 100 filters here. 

\begin{figure}[t]	
\centering
\centerline{\includegraphics[scale=0.2]{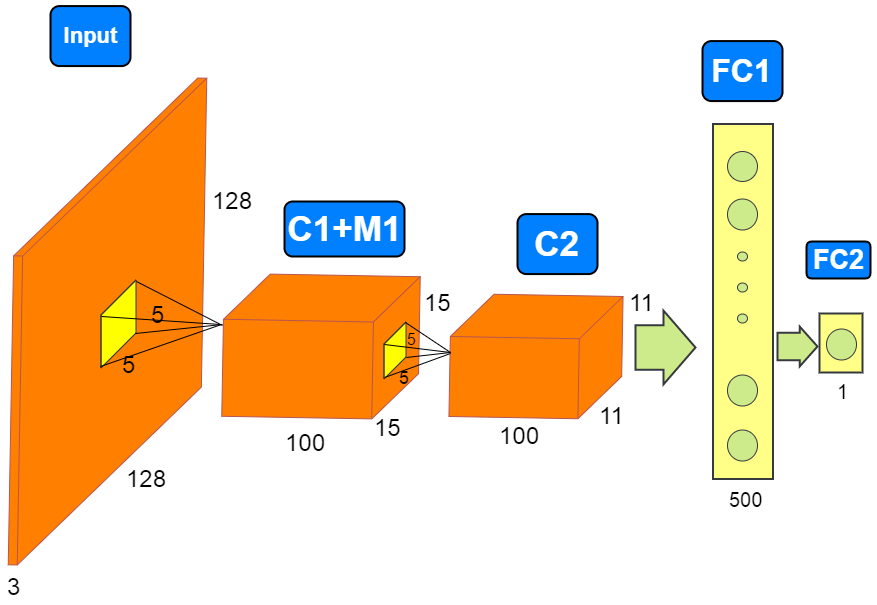}}
\caption{The architecture of the CNN designed for face detection in HEVC feature images. }
 \label{fig:CNN}
 	 \end{figure}

The final CNN architecture is shown in Fig.~\ref{fig:CNN}, where ``C'' indicates convolutional layer, ``M'' indicates max-pooling layer and ``FC'' indicates fully-connected layer. Rectified linear unit (ReLU) functions are used for activation in convolutional layers and sigmoid is used in the output layer. A dropout of rate 0.25 is used in FC1. The same architecture is used for $64 \times 64$ inputs, but the input layer in this case is $64 \times 64 \times 3$.

\section{Experimental Results}
\label{sec:Exp_results}

For the experiments, 13,000 face images were taken from the Labeled Faces in the Wild (LFW) dataset~\cite{LFW,LFW_aligned}, and 120,000 non-face images were taken from the Large Scale Visual Recognition Challenge~\cite{ILSVRC}. 
From these images, 15\% was used for testing purposes and the remaining 85\% (i.e., around 11k face images and 102k non-face images) were used for training. Note that the number of negative (non-face) samples is around 9 times larger than the number of positive (face) samples in the dataset. The reason is that in practice, a face detector is likely to see non-faces much more often than faces, which should be reflected in the training.    


  

Experiments were run on a desktop machine with Ubuntu 16.04, 128 GB RAM, Intel i7 processor at 3.6 GHz and Nvidia Titan X GPU. Image patches of size $64 \times 64$ and $128 \times 128$  were HEVC intra-coded with QP $\in \{22, 32, 42\}$ using~\cite{HM}. A separate CNN model with the same architecture shown in Fig.~\ref{fig:CNN} was trained for each of the six combinations of image size and QP. In each case, the training was carried on until convergence, where convergence is defined as the event of validation loss not decreasing for three consecutive epochs. 


The performance of the trained CNN models based on compression-domain data is shown in the first column of Table~\ref{Tbl:results_pr}, Table~\ref{Tbl:results_recall} and Table~\ref{Tbl:results_F1} in terms of  Precision, Recall and F1-measure. These are computed from True Positives (TP), False Positives (FP), True Negatives (TN) and False Negatives (FN) as Precision $= \frac{TP}{TP+FP}$, Recall $= \frac{TP}{TP+FN}$ and F1 $= 2 \frac{\text{Presicion} \cdot \text{Recall}}{\text{Presicion} + \text{Recall}}$. As seen in the tables, very high Precision, Recall and F1-measure are achieved for QP=32 (indicated in bold). One reason is the CNN architecture in Fig.~\ref{fig:CNN} was selected based on the data obtained with QP=32. However, the same architecture works reasonably well for other QPs, although improvements could be expected by developing separate architectures for each QP. 

With QP=42, the encoder tends to choose lager PUs, which reduces the amount of detail in face regions of our feature images (Fig.~\ref{fig:QP_effect}). Hence, it becomes more difficult to distinguish faces from non-faces based on feature images, so the accuracy drops. With QP=22, small PUs are more frequently selected in both face and non-face patches. Hence, in this case also the accuracy drops, although not as much as with QP=42. The results also show that for each QP, higher accuracy is obtained for $128 \times 128$ feature images compared to the $64 \times 64$ case.


\begin{table}[b]
	\centering
	\caption{Precision of the proposed face detection method using compression-domain data compared to the method proposed in \cite{benchmark} based on the pixel-domain data }
	\label{Tbl:results_pr}
	\begin{tabular}{|c|c|c|c|c|}
		\hline
		Size, QP             & Proposed method           &  Method in \cite{benchmark} \\ \hline \hline
		$64 \times 64$, 22   &    0.86                   &    0.89                        \\ \hline
		$128 \times 128$, 22 &    0.95                   &    0.94                        \\ \hline
		$64 \times 64$, 32   &    \bfseries 0.97         &    0.90                        \\ \hline
		$128 \times 128$, 32 &    \bfseries 0.99         &    0.95                        \\ \hline
		$64 \times 64$, 42   &    0.79                   &    0.90                        \\ \hline
		$128 \times 128$, 42 &    0.89                   &    0.94                        \\ \hline
		\hline
		Average              &    0.91                   &    0.92                         \\ \hline		
	\end{tabular}
\end{table}

\begin{table}[t]
	\centering
	\caption{Recall of the proposed face detection method using compression-domain data compared to the method proposed in \cite{benchmark} based on the pixel-domain data}
	\label{Tbl:results_recall}
	\begin{tabular}{|c|c|c|c|c|}
		\hline
		Size, QP             & Proposed method           &  Method in \cite{benchmark} \\ \hline \hline
		$64 \times 64$, 22   &    0.72                   &    0.70                        \\ \hline
		$128 \times 128$, 22 &    0.88                   &    0.87                        \\ \hline
		$64 \times 64$, 32   &    \bfseries 0.95         &    0.72                        \\ \hline
		$128 \times 128$, 32 &    \bfseries 0.98         &    0.85                        \\ \hline
		$64 \times 64$, 42   &    0.50                   &    0.64                        \\ \hline
		$128 \times 128$, 42 &    0.79                   &    0.85                        \\ \hline
		\hline
		Average              &    0.80                   &    0.77                        \\ \hline		
	\end{tabular}
\end{table}

\begin{table}[t]
	\centering
	\caption{F1-measure of the proposed face detection method using compression-domain data compared to the method proposed in \cite{benchmark} based on the pixel-domain data}
	\label{Tbl:results_F1}
	\begin{tabular}{|c|c|c|c|c|}
		\hline
		Size, QP             & Proposed method           &  Method in \cite{benchmark} \\ \hline \hline
		$64 \times 64$, 22   &    0.79                   &    0.78                        \\ \hline
		$128 \times 128$, 22 &    0.91                   &    0.90                        \\ \hline
		$64 \times 64$, 32   &    \bfseries  0.96        &    0.80                        \\ \hline
		$128 \times 128$, 32 &    \bfseries 0.99         &    0.89                        \\ \hline
		$64 \times 64$, 42   &    0.61                   &    0.75                       \\ \hline
		$128 \times 128$, 42 &    0.84                   &    0.89                        \\ \hline
		\hline
		Average              &    0.85                  &     0.84                      \\ \hline		
	\end{tabular}
\end{table}



\begin{figure}[t]	
	\centering
	\centerline{\includegraphics[scale=0.2]{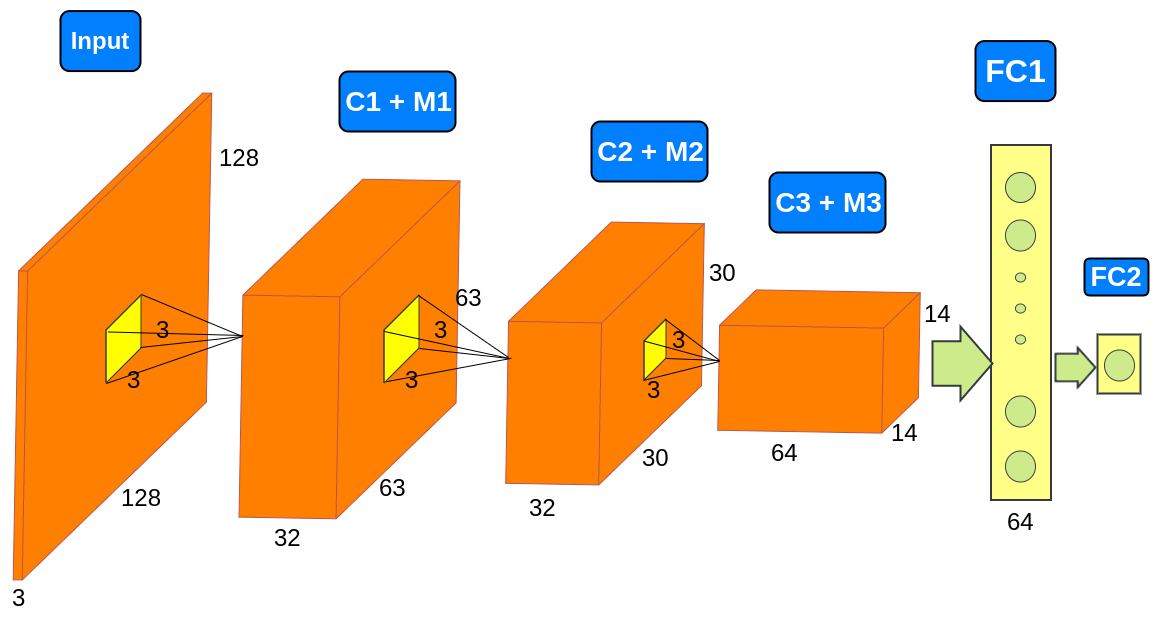}}
	\caption{The architecture of the face detection CNN in~\cite{benchmark}.}
	\label{fig:bnch_CNN}
\end{figure}

Next, we examine how the accuracy of the proposed method compares with conventional pixel-domain face detection (Fig.~\ref{fig:flow_chart}(a)). To this end, we selected as a benchmark a recent CNN-based face detector~\cite{benchmark}, whose architecture is shown in Fig.~\ref{fig:bnch_CNN} for a $128 \times 128$ input. We trained a separate CNN on the fully-decoded image patches from our dataset for the six combinations of size and QP. The precision, recall and F1-measure results are shown in the second column of Table~\ref{Tbl:results_pr}, Table~\ref{Tbl:results_recall} and Table~\ref{Tbl:results_F1}, respectively.
  

Comparison of the results in columns of Table~\ref{Tbl:results_pr}, Table~\ref{Tbl:results_recall} and Table~\ref{Tbl:results_F1} shows that the accuracies of the two approaches are comparable. The results of the model from~\cite{benchmark} are somewhat more consistent; for example, the recall values do not vary from case to case as much as those of our proposed method, which may be expected on account of variation in feature images for different QPs (Fig.~\ref{fig:QP_effect}). But the averages across all cases are fairly similar, indicating that face detection from HEVC bitstreams is indeed possible with accuracy comparable to conventional pixel-domain detectors. 

Note that the architecture of the CNN model in~\cite{benchmark} was selected based on a different dataset. Hence, the performance of pixel-domain face detection could be expected to be higher had the architecture been chosen based on our data. However, the same could be said about our model, which was selected for QP=32 (the case where it achieved very high accuracy) - improvements can be expected by tailoring the architecture to each QP. 

We have already seen that the required decoding time is significantly reduced if HEVC features images are used for face detection compared to fully-reconstructed images. But the savings do not end there. As seen from Figs.~\ref{fig:CNN} and~\ref{fig:bnch_CNN}, the proposed CNN architecture is shallower than the one from~\cite{benchmark}. Specifically, the proposed CNN comprises two convolutional layers, one max-pooling layer, and two fully-connected layers, while the CNN in~\cite{benchmark} uses three convolutional layers, three max-pooling layers and two fully-connected layers. 

Note that the overall number of parameters is higher in the proposed CNN (Table~\ref{Tbl:CNN_complexity_num}), due to the fact that it uses more filters (200 in total, compared to 128 in~\cite{benchmark}) and has a larger number of nodes in the fully-connected layer (500, compared to 64 in~\cite{benchmark}). This makes the training of our CNN longer, but that is an off-line operation. Once trained, the proposed CNN runs 2-3 times faster compared to the one from~\cite{benchmark}, because it is shallower. The run-time results for the two models measured on our system are shown in Table~\ref{Tbl:CNN_complexity_time}. 


\begin{table}[t]
\centering
\caption{Number of model parameters}
\label{Tbl:CNN_complexity_num}
\begin{tabular}{|c|c|c|}
\hline
Input size & \begin{tabular}[c]{@{}c@{}}Number of model \\ parameters in Fig.~\ref{fig:CNN} \end{tabular} & \begin{tabular}[c]{@{}c@{}}Number of model \\parameters in Fig.~\ref{fig:bnch_CNN} \end{tabular} \\ \hline
64 $\times$ 64    & 708,701                                                                         & 176,225                                                                         \\ \hline
128 $\times$ 128  & 6,308,701                                                                        & 831,585                                                                         \\ \hline
\end{tabular}
\end{table}

\begin{table}[t]
\centering
\caption{Evaluation time per input patch averaged over all the test images and QP values}
\label{Tbl:CNN_complexity_time}
\begin{tabular}{|c|c|c|}
\hline
Input size & \begin{tabular}[c]{@{}c@{}}Evaluation time (s)\\ for the model in Fig.~\ref{fig:CNN} \end{tabular} & \begin{tabular}[c]{@{}c@{}}Evaluation time (s) \\ for the model in Fig.~\ref{fig:bnch_CNN} \end{tabular} \\ \hline
64 $\times$ 64    & 6 $\times 10^{-4}$                                              & $ 20 \times$$10^{-4}$                                                  \\ \hline
128 $\times$ 128  & 35$\times 10^{-4}$                                              & $ 82 \times 10^{-4}$                                                   \\ \hline
\end{tabular}
\end{table}

\section{Conclusion and Future Work}
\label{sec:Conclusion}
This paper proposed a face detection method based on features derived from partially-decoded HEVC bitstreams. In the proposed method, feature images are created from the output of the HEVC entropy decoder and then fed to a CNN that determines whether or not the input patch contains a face. The experimental results show that the proposed method achieves high detection accuracy, comparable to CNN-based pixel-domain face detection. The results also showed that HEVC feature images can change considerably when QP changes, which suggests that a separate CNN model could be trained for various QPs.



The proposed face detector is suitable for face localization in large images using a sliding-window approach. However, a more recent generation of object detectors~\cite{SSD, Yolo9000} avoids the use of sliding windows by testing only a smaller group of object candidates. Our goal in future work is to extend the proposed face detector to enable fast face localization in large images using a similar approach.




\bibliographystyle{IEEEtran}

\bibliography{references}

\begin{thebibliography}{10}
\providecommand{\url}[1]{#1}
\csname url@samestyle\endcsname
\providecommand{\newblock}{\relax}
\providecommand{\bibinfo}[2]{#2}
\providecommand{\BIBentrySTDinterwordspacing}{\spaceskip=0pt\relax}
\providecommand{\BIBentryALTinterwordstretchfactor}{4}
\providecommand{\BIBentryALTinterwordspacing}{\spaceskip=\fontdimen2\font plus
\BIBentryALTinterwordstretchfactor\fontdimen3\font minus
  \fontdimen4\font\relax}
\providecommand{\BIBforeignlanguage}[2]{{%
\expandafter\ifx\csname l@#1\endcsname\relax
\typeout{** WARNING: IEEEtran.bst: No hyphenation pattern has been}%
\typeout{** loaded for the language `#1'. Using the pattern for}%
\typeout{** the default language instead.}%
\else
\language=\csname l@#1\endcsname
\fi
#2}}
\providecommand{\BIBdecl}{\relax}
\BIBdecl

\bibitem{Kay_Detection}
S.~M. Kay, \emph{Fundamentals of Statistical Signal Processing: Detection
  Theory}.\hskip 1em plus 0.5em minus 0.4em\relax Prentice Hall, 1998, vol.~II.

\bibitem{cascade}
H.~Li, Z.~Lin, X.~Shen, J.~Brandt, and G.~Hua, ``A convolutional neural network
  cascade for face detection,'' in \emph{Proc. IEEE CVPR'15}, 2015, pp.
  5325--5334.

\bibitem{facial_parts}
S.~Yang, P.~Luo, C.-C. Loy, and X.~Tang, ``From facial parts responses to face
  detection: A deep learning approach,'' in \emph{Proc. IEEE ICCV'15}, 2015,
  pp. 3676--3684.

\bibitem{hyperface}
R.~Ranjan, V.~M. Patel, and R.~Chellappa, ``Hyperface: A deep multi-task
  learning framework for face detection, landmark localization, pose
  estimation, and gender recognition,'' \emph{arXiv:1603.01249}, 2016.

\bibitem{He2014_arXiv}
K.~{He} and J.~{Sun}, ``{Convolutional Neural Networks at Constrained Time
  Cost},'' \emph{arXiv:1412.1710}, 2014.

\bibitem{HEVC}
G.~J. Sullivan, J.~Ohm, W.-J. Han, and T.~Wiegand, ``Overview of the high
  efficiency video coding ({HEVC}) standard,'' \emph{IEEE Trans. Circuits Syst.
  Video Technol.}, vol.~22, no.~12, pp. 1649--1668, 2012.

\bibitem{hevc_ctc}
F.~Bossen, ``Common {HM} test conditions and software reference
  configurations,'' in \emph{ISO/IEC JTC1/SC29 WG11 {m28412}, {JCTVC-L1100}},
  Jan. 2013.

\bibitem{hevc_counting}
Y.~W. Chen, K.~Chen, S.~Y. Yuan, and S.~Y. Kuo, ``Moving object counting using
  a tripwire in {H.265/HEVC} bitstreams for video surveillance,'' \emph{IEEE
  Access}, vol.~4, pp. 2529--2541, 2016.

\bibitem{moving_object_hevc}
L.~Zhao, Z.~He, W.~Cao, and D.~Zhao, ``Real-time moving object segmentation and
  classification from {HEVC} compressed surveillance video,'' \emph{IEEE Trans.
  Circuits Syst. Video Technol.}, to appear.

\bibitem{kb_tip_2013}
S.~H. Khatoonabadi and I.~V. Bajic, ``Video object tracking in the compressed
  domain using spatio-temporal markov random fields,'' \emph{IEEE Trans. Image
  Processing}, vol.~22, no.~1, pp. 300--313, Jan. 2013.

\bibitem{LFW}
G.~B. Huang, M.~Ramesh, T.~Berg, and E.~Learned-Miller, ``Labeled faces in the
  wild: A database for studying face recognition in unconstrained
  environments,'' Technical Report 07-49, University of Massachusetts, Amherst,
  Tech. Rep., 2007.

\bibitem{LFW_aligned}
G.~B. Huang, M.~Mattar, H.~Lee, and E.~Learned-Miller, ``Learning to align from
  scratch,'' in \emph{NIPS}, 2012.

\bibitem{ILSVRC}
O.~Russakovsky, J.~Deng, H.~Su, J.~Krause, S.~Satheesh, S.~Ma, Z.~Huang,
  A.~Karpathy, A.~Khosla, M.~Bernstein, A.~C. Berg, and L.~Fei-Fei, ``{ImageNet
  Large Scale Visual Recognition Challenge},'' \emph{Int. J. Computer Vision},
  vol. 115, no.~3, pp. 211--252, 2015.

\bibitem{HM}
``{HEVC} reference software ({HM} 16.5),''
  \url{https://hevc.hhi.fraunhofer.de/trac/hevc/browser/tags/HM-16.15},
  accessed: 2017-05-27.

\bibitem{benchmark}
J.~Duan, S.~Liao, S.~Zhou, and S.~Z. Li, ``Face classification: A specialized
  benchmark study,'' in \emph{Chinese Conference on Biometric
  Recognition}.\hskip 1em plus 0.5em minus 0.4em\relax Springer, 2016, pp.
  22--29.

\bibitem{SSD}
W.~Liu, D.~Anguelov, D.~Erhan, C.~Szegedy, S.~Reed, C.-Y. Fu, and A.~C. Berg,
  ``{SSD:} single shot multibox detector,'' in \emph{Proc. ECCV}.\hskip 1em
  plus 0.5em minus 0.4em\relax Springer, 2016.

\bibitem{Yolo9000}
J.~{Redmon} and A.~{Farhadi}, ``{{YOLO9000:} Better, Faster, Stronger},''
  \emph{arXiv:1612.08242}, Dec. 2016.

\end{thebibliography}

\end{document}